\def\year{2020}\relax
\documentclass[letterpaper]{article} 
\usepackage{aaai20}  
\usepackage{times}  
\usepackage{helvet}  
\usepackage{courier}  
\usepackage[hyphens]{url}  
\usepackage{graphicx}  
\usepackage[utf8]{inputenc}

\usepackage{amsmath}
\usepackage{booktabs}
\usepackage{algorithm}
\usepackage{algorithmic}

\usepackage{mathtools}
\usepackage{amssymb}
\usepackage{amsthm}

\newtheorem{proposition}{Proposition}

\newtheorem{definition}{Definition}

\DeclareMathOperator*{\argmin}{arg\,min}
\DeclareMathOperator*{\argmax}{arg\,max}
\usepackage{scalerel,stackengine}
\stackMath
\newcommand\reallywidehat[1]{%
\savestack{\tmpbox}{\stretchto{%
  \scaleto{%
    \scalerel*[\widthof{\ensuremath{#1}}]{\kern-.6pt\bigwedge\kern-.6pt}%
    {\rule[-\textheight/2]{1ex}{\textheight}}
  }{\textheight}%
}{0.5ex}}%
\stackon[1pt]{#1}{\tmpbox}%
}

\title{Reinforcement Learning from Imperfect Demonstrations \\ under Soft Expert Guidance}
\author{
Mingxuan Jing\thanks{These two authors contributed equally. Correspondence to Fuchun Sun.}$^1$,
Xiaojian Ma\footnotemark[1]$^1$$^2$,
Wenbing Huang$^1$,
Fuchun Sun$^1$,
Chao Yang$^1$,
Bin Fang$^1$,
Huaping Liu$^1$
\\
$^1$Beijing National Research Center for Information Science and Technology (BNRist),\\ State Key Lab on Intelligent Technology and Systems,\\ Department of Computer Science and Technology, Tsinghua University, Beijing 100084, China\\
$^2$Center for Vision, Cognition, Learning and Autonomy, Department of Computer Science, UCLA, CA 90095, USA\\ 
\texttt{jmx16@mails.tsinghua.edu.cn, maxiaojian@ucla.edu}\\ \texttt{fcsun@tsinghua.edu.cn, hwenbing@126.com}
}

\begin{document}
%
\maketitle
\begin{abstract}
In this paper, we study \emph{Reinforcement Learning from Demonstrations (RLfD)} that improves the exploration efficiency of Reinforcement Learning (RL) by providing expert demonstrations. Most of existing RLfD methods require demonstrations to be perfect and sufficient, which yet is unrealistic to meet in practice. To work on imperfect demonstrations, we first define an imperfect expert setting for RLfD in a formal way, and then point out that previous methods suffer from two issues in terms of optimality and convergence, respectively. Upon the theoretical findings we have derived, we tackle these two issues by regarding the expert guidance as a soft constraint on regulating the policy exploration of the agent, which eventually leads to a constrained optimization problem. We further demonstrate that such problem is able to be addressed efficiently by performing a local linear search on its dual form.
Considerable empirical evaluations on a comprehensive collection of benchmarks indicate our method attains consistent improvement over other RLfD counterparts.
\end{abstract}

\section{Introduction}\label{sec:intro}
Reinforcement Learning (RL)~\cite{rl_book} enables robots to acquire skills by interacting with the environment.
Despite the conspicuous advancements they have attained, typical RL methods suffer from the exploration issue that performing exploration over novel action-state trajectories is inefficient, and is not spontaneously guaranteed when the reward signals are sparse or incomplete. Thus, a fairly of approaches~\cite{ijcaipofd1,ijcaipofd2,ijcaipofd3,pofd,thor} have resorted to the combination of RL with expert demonstrations (containing action-state trajectories), giving rise to a new research vein that exploits expert demonstrations to help policy exploration of the agent. We refer this vein as Reinforcement Learning from Demonstrations (RLfD) in this paper. 

\begin{figure}[t!]
\begin{center}
\centering
\includegraphics[width=1.\linewidth]{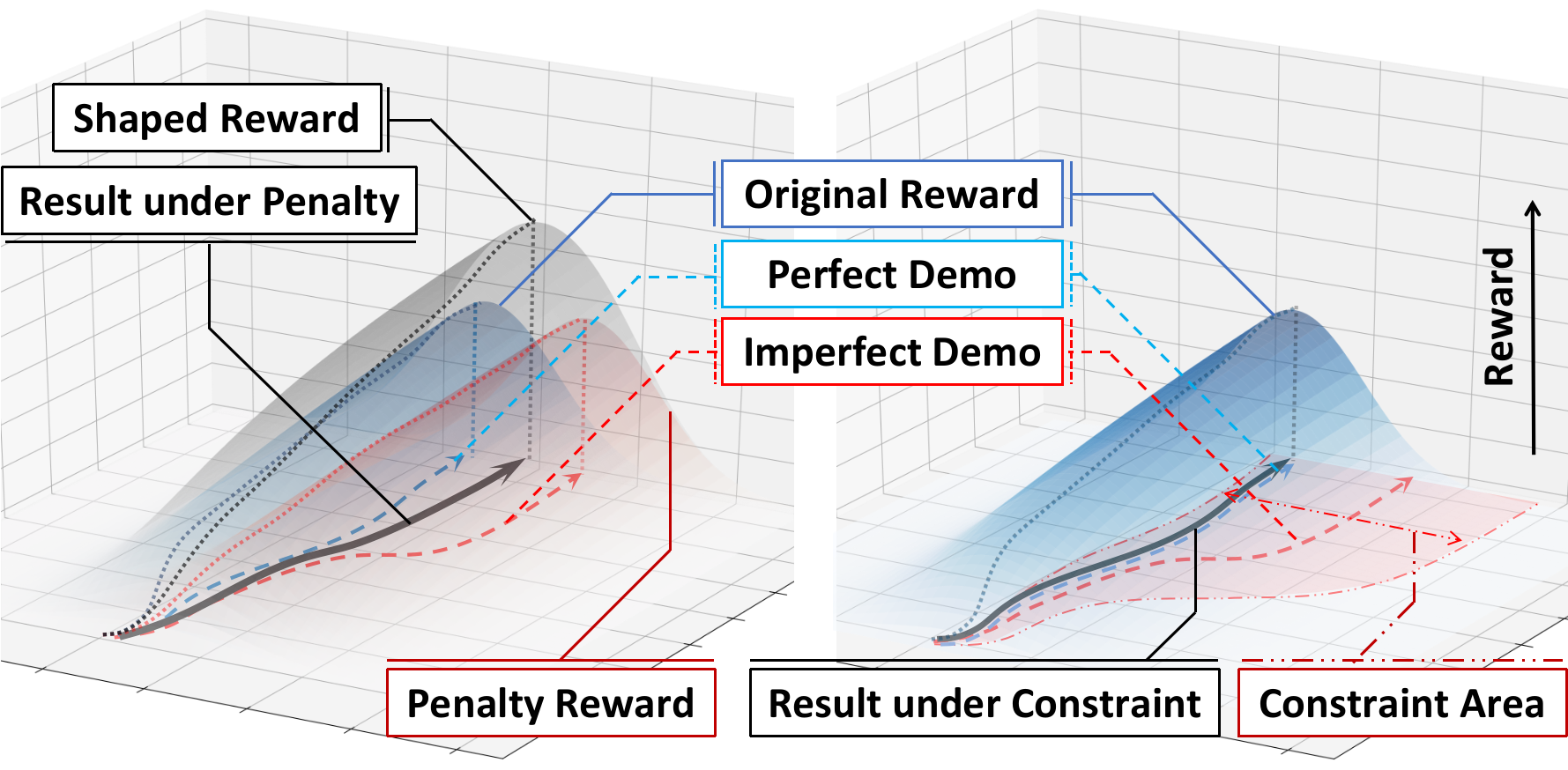}
\caption{An overview of our RLfD method using soft constraint versus existing approaches using penalty departures. \textbf{Left}: In penalty method, agent seeks to maximize the shaped reward which may induce non-optimal solution. \textbf{Right}: Proposed soft constraint will guide the agent to explore towards areas with high reward without altering the optimality.}
\label{fig:overview}
\end{center}
\end{figure}

Early RLfD methods enhance RL by either putting expert demonstrations into a replay buffer for value estimation~\cite{dqfd,ddpgfd} or applying them to pre-train the policy in a supervised manner~\cite{pretrain1,pretrain2}, both of which, however, simply regard demonstrations as data-augmentations without making full use of them during the policy optimization procedure. To address this weakness, modern RLfD approaches~\cite{thor,pofd} absorb ingredients from Imitation Learning~\cite{lfd1,Apprenticeship,MaxEntIRL,GAIL,pbganirl,iddm}, and encourage the agent to mimic the demonstrated behaviors when the environmental feedback is rare or even unavailable. Specifically, they reshape the native reward in RL by adding a demonstration-guided term to force expert-alike exploration. 

Whereas encouraging expert-alike actions does help in avoiding futile exploration, continuously enforcing such type of rewards during the whole learning phase is problematic if the provided demonstrations are imperfect. Here, the notion of \emph{imperfect demonstrations} implies two senses: I. The quality of demonstrations is imperfect, which could be caused by data collection noise or intrinsically produced by the immaturity of the expert. II. The number of demonstrations is insufficient, which is due to the consuming resource and effort in collection. The imperfectness of demonstrations will potentially, if not always, make the convergence of the agent policy to be sub-optimal. As illustrated in Figure~\ref{fig:overview} and non-strictly speaking, the learned agent policy by existing RLfD methods will converge to a point nearby the underlying expert policy. If the demonstrations/expert policies are imperfect, we have no guarantee to obtain better agent policy (or even have a potential detriment to the policy searching) by always minimizing its divergence to the expert behavior.    

In this paper, we propose to conduct RL from imperfect demonstrations by applying expert guidance in a \emph{soft} way. To illustrate our idea, let us revisit the example in Figure~\ref{fig:overview}. We assume that the optimal agent policy still locates within a certain region around the imperfect expert policy (denoted by the red area in the \textbf{Right} of  Figure~\ref{fig:overview}), and once the agent policy is within this region, its optimization is only affected by the interaction with the environment and is no longer influenced by demonstrations. The intuition behind is that the expert demonstrations---even when they are imperfect---are able to characterize what actions are good in general but not precisely. Conventional RLfD methods fix the demonstration reward during the whole learning procedure and are not flexible enough to meet our requirement. 

Towards our purpose, we reformulate the RLfD task as a constrained policy optimization problem~\cite{cmdp,cpo,rcpo}, where the goal is formulated by the native RL objective and the constraint is to bound the exploration region around the demonstrations under a certain threshold. By this formulation, the expert demonstrations regulate the agent policy updating only when the policy is outside the constraint region, which is consistent with our assumption mentioned above. Nevertheless, solving the constrained optimization problem is non-trivial. To tackle it effectively, we propose to search the optimal policy update for each step with a linearized sub-objective. Through leveraging its dual form, we can significantly reduce the problem size and empowers the scalability to policy models with high-dimensional parameter space like neural networks. We provide more details in Sec.~\ref{sec:method}.

We summarize our contributions as follows.
\begin{itemize}
    \item To the best of our knowledge, we are the first to formulate RLfD as a constrained optimization problem, by which we are able to make full use of imperfect demonstrations in a soft and also more effective manner.
    \item We develop an efficient method to solve the proposed constrained optimization problem with scalable policy model like deep neural networks.
    \item With imperfect demonstrations, our method achieves consistent improvement over other RLfD counterparts on several challenging physical control benchmarks.
\end{itemize}

The rest of paper is organized as follows: In Sec.~\ref{sec:bg}, we first provide necessary notations and preliminaries about the subject of RLfD. Then our proposed method will be detailed in Sec.~\ref{sec:method} with analysis and efficient implementation. The discussion on some related research will be included in Sec.~\ref{sec:discuss}. Finally, experimental evaluations will be demonstrated in Sec.~\ref{sec:exp}.

\section{Preliminaries}\label{sec:bg}
\noindent
\textbf{Notations.} For modeling the action decision process in our context, a standard Markov decision process (MDP)~\cite{rl_book} $(\mathcal{S}, \mathcal{A}, r, \mathcal{T}, \mu, \gamma)$ is considered, where $\mathcal{S}$ and $\mathcal{A}$ denotes the space of feasible states and actions respectively, $r(s,a)\rightarrow \mathbb{R}$ is the reward function,
$\mathcal{T}(s'|s,a)$ and $\mu(s)$ represent the transition probability and initial state distribution and $\gamma \in (0, 1)$ is the discount factor. A stochastic policy $\pi(a|s): \mathcal{S}\times\mathcal{A} \rightarrow [0,1]$ maps state into action distribution. A trajectory $\zeta$ is given by the sequence of state-action pairs $\{(s_0, a_0),(s_1, a_1),...\}$. 

\noindent
\textbf{Occupancy measure.} The concept of occupancy measure~\cite{mdp_book,irl_lp} defined below characterizes the distribution of the state-action pairs within the exploration trajectories when policy $\pi$ is executed, which will be useful in the following analysis.
\begin{definition}[Occupancy Measure]\label{def:rho_sa}
Given a stationary policy $\pi$, let $\rho_\pi(s): \mathcal{S} \rightarrow \mathbb{R}$ and $\rho_\pi(s, a): \mathcal{S}\times\mathcal{A} \rightarrow \mathbb{R}$ denote the density of the state distribution and the joint distribution for state and action under the policy $\pi$,
\begin{align}\label{equ:rho_s}
\begin{split}
    &\rho_\pi(s)  \triangleq \sum_{t=0}^{\infty}\gamma^t P(s_t=s|\pi) \\
    &\rho_\pi(s,a) \triangleq \rho_\pi(s)\pi(a|s)\text{.}
\end{split}
\end{align}
We name $\rho_\pi(s,a)$ as occupancy measure of policy $\pi$. 
\end{definition}
\noindent
\textbf{Formulation of RLfD.} The objective of RL is to maximize the cumulative expected (discounted) return along the whole decision procedure $\eta(\pi) = \mathbb{E}_\pi[\sum_t^\infty \gamma^tr(s_t, a_t)]$, given current action policy $\pi$~\cite{rl_book}. While RLfD enhances RL with providing a set of demonstrated trajectories $\mathcal{D} = \{\zeta_0, \cdots\}$ draw from a referred expert with policy $\pi_E$ as an extra guidance other than reward. Such expert data can be useful notably when the environmental feedback is sparse or delayed~\cite{ICM}, in which the agent may suffer from ineffective explorations since positive feedback could rarely occur. 

\noindent
\textbf{RLfD with penalty departures.} Some previous research\cite{ijcaipofd1,pofd} suggest exploring towards the area that frequently visited by expert policy $\pi_E$, because it may provide a higher and denser return that agent can benefit from. As it mentioned above, such visiting frequency is essentially characterized by expert's occupancy measure. Intuitively, we can leverage the distribution discrepancy between the occupancy measure of expert and agent as an extra feedback signal to encourage this exploration behavior, which gives us the following objective
\begin{align}\label{equ:penalty}
    \min_\pi \mathcal{L}_\pi = -\eta(\pi) + \lambda\cdot \mathbb{D}\left[\rho_\pi(s,a) \| \rho_E(s,a)\right]\text{,}
\end{align}
where $\mathbb{D}(\cdot\|\cdot)$ and $\rho_E(s,a)$ depict any discrepancy measure and expert's occupancy measure respectively, $\lambda$ is an adjustable weight. We refer~\eqref{equ:penalty} as \emph{RLfD with penalty departures} in the following context since the discrepancy is introduced as a penalty function upon the original objective of RL and can be approximated through expert demonstrations. 

\section{Methodology}\label{sec:method}
In this section, we will first introduce the new setting of \emph{imperfect} expert for RLfD and emphasize the optimality and convergence issues in the penalty method, which essentially motivates our approach to employ expert guidance as a soft constraint instead. We also demonstrate that such constrained optimization problem can be solved efficiently by performing a local linear search on its dual form, maintaining its scalability to complex policy model like deep neural networks. Finally, we provide a practical implementation of our method.

\subsection{Expert Guidance as a Soft Constraint: Towards RLfD with an Imperfect Expert}\label{sec:soft_constraint}
We now illustrate the imperfect setting for RLfD. As it mentioned in Sec.~\ref{sec:intro}, the imperfectness here is raised from two facets: \textbf{quality} and \textbf{amount} of available demonstrations. Here we first focus on the quality, and the issue on the amount of demonstrations will be discussed later. Compared to the perfect expert setting that assumed the expert policy has already maximized the expected return~\cite{ijcaipofd1,pofd}, an imperfect expert employs a policy that still not converge to an expected local optimum \emph{w.r.t.} the considered RL objective. Without
loss of generality, an imperfect expert, can be defined as follows.
\begin{definition}[Imperfect Expert Policy]\label{def:imperfect}
Denoting $\pi_{\theta^{+}}$ and $\pi_{\theta^{-}}$ as perfect and imperfect expert policies respectively. $\pi_{\theta^{-}}$ either attains local optimum with a lower return than $\pi_{\theta^{+}}$ or does not belong to any local optima.
\begin{align}
    &\pi_{\theta^{+}} \in \left\{\pi: \argmax_\pi \eta(\pi)\texttt{\ AND\ }\frac{\partial \eta(\pi_{\theta^{+}})}{\partial\theta^+} = 0\right\}  \nonumber\\
    &\begin{array}{ @{} r l @{} }
    \pi_{\theta^{-}} \in \bigg\{\pi : \left\{\frac{\partial\eta(\pi_{\theta^{-}})}{\partial\theta^-} = 0\texttt{\ AND\ }\eta(\pi_{\theta^{-}}) < \eta(\pi_{\theta^{+}})\right\} \\
    \texttt{\ OR\ }\left\{\frac{\partial\eta(\pi_{\theta^{-}})}{\partial\theta^-} \neq 0\right\}\bigg\}\end{array}\text{,}\nonumber
\end{align}
where $\eta(\pi)$ is the objective of currently considered RL task.
\end{definition}
The penalty method presented in Sec.~\ref{sec:bg} works comparably well when expert is optimal~\cite{ijcaipofd1,pofd}. However, optimizing the composite sum of two parts in~\eqref{equ:penalty} under imperfect expert setting is problematic, as it may alter the optimality and induces no convergence guarantee for the original RL objective. The following propositions illustrate this issue formally.
\begin{proposition}\label{prop:1}
Denoting $\pi_{\theta^\star} = \argmax_\pi \eta(\pi)$ as the optimal policy under the given RL objective $\eta(\pi)$. Then for the additional distribution discrepancy term $\mathbb{D}\left[\rho_\pi\|\rho_E\right]$ in~\eqref{equ:penalty}, $\frac{\partial\mathbb{D}[\rho_{\pi_{\theta}}\| \rho_{\pi_{\theta^{+}}}]}{\partial\theta}\big|_{\theta=\theta^\star}= 0$. But when an imperfect expert $\pi_{\theta^-}$ is adopted, this result does not hold under certain conditions.
\end{proposition}
This proposition presents an intuitive result that the optimal policy for a given RL task can't always be an optimum of the additional discrepancy term in~\eqref{equ:penalty} under imperfect demonstrations. We will further show that this may make~\eqref{equ:penalty} converge to a solution that is non-optimal for the original RL problem.
\begin{proposition}\label{prop:2}
When the penalty method~\eqref{equ:penalty} under imperfect demonstrations converges to a local optimum $\pi_\theta$, it can't always be the optimal solution for the original RL objective.
\begin{align}
        \frac{\partial\mathcal{L}_{\pi_\theta}}{\partial\theta} = 0 \nRightarrow \pi_\theta = \argmax_\pi \eta(\pi)\text{.}\nonumber
\end{align}
While under the same certain condition as Proposition~\ref{prop:1}, we can obtain an even stronger conclusion
\begin{align}
    \frac{\partial\mathcal{L}_{\pi_\theta}}{\partial\theta} = 0 \Rightarrow \eta(\pi_\theta) < \max_\pi \eta(\pi)\text{.}\nonumber
\end{align}
\end{proposition}
The two propositions above imply that, under the imperfect setting, the additional penalty term will substantially change the optimization landscape of original RL problem and may induce convergence to a non-optimal solution. Although it can offer positive guidance in the early training phase, it will soon become misleading and prevent the policy from attaining higher return. To tackle this issue, we propose to transform the distribution discrepancy penalty term into a constraint instead. This intuition is actually based on the following observation:
\begin{proposition}\label{prop:3}
There exists a bounded tolerance factor $d$ such that the optimal policy $\pi_{\theta^\star}$ always stay within an area closer to the demonstrations specified by $d$, even when the demonstrations are drawn from an imperfect expert.
\begin{align}
        \exists d\in [0, \infty), \mathbb{D}\left[\rho_{\pi_{\theta^\star}}\|\rho_{\pi_{\theta^-}}\right] \leqslant d, \pi_{\theta^\star} = \argmax_\pi \eta(\pi) \text{.}\nonumber
\end{align}
\end{proposition}
From the perspective of optimization, it suggests that using constraint could better fit the imperfect expert setting by two reasons. \textbf{1. Optimality.} Refer to Proposition~\ref{prop:3}, 
given a proper tolerance $d$, once the optimal policy satisfies the constraint, the new constrained optimization problem will share the same optimal solution with the original RL problem. \textbf{2. Convergence.} The constraint only affects policy update when it is not satisfied. Therefore, when the policy improves to a certain extent, \emph{i.e.} stays within the constraint, it will only learn from the original reward feedback and finally converge to the optimality of the original RL problem.
As a conclusion, compared to the penalty method, the constraint method can leverage the imperfect demonstrations for guiding the policy while eliminating their side effects in optimization, thus can work better with imperfect experts.

For another facet of imperfectness, \emph{i.e.} amount, as the expert data is mainly introduced for computing the distribution discrepancy in our context, the issue of insufficient amount of demonstrations will essentially rely on the estimation error to the discrepancy, which may induce bias to policy update especially when the gradient step is relatively large. We refer to the idea of local policy search~\cite{lps,lps2} to alleviate this issue by making conservative gradient step instead with an auxiliary constraint on the change of Kullback-Leibler (KL) divergence of the updated policy. 

\noindent
\textbf{Overall optimization problem.} By combining the discrepancy constraint and local policy search, the eventual optimization problem ($k$-th step) with imperfect expert $\pi_{\theta^{-}}$ is
\begin{align}
\begin{split}\label{equ:final_objective}
    \theta_{k+1} = \argmax_\theta~~& \eta(\pi_{\theta_k}) \\
    \text{s.t.}~~&\mathbb{D}\left[\rho_{\pi_{\theta_k}}(s, a) \| \rho_{\pi_{\theta^{-}}}(s, a)\right] \leqslant d_k \\
    &\mathbb{D}_{\operatorname{KL}}\left[\pi_{\theta_k}(a|s) \| \pi_{\theta_{k+1}}(a|s)\right] \leqslant \delta\text{,}
\end{split}
\end{align}
where $\delta$ is the tolerance of the KL constraint. The remaining issue now is how to determine the tolerance factor $d_k$ for the discrepancy constraint in each step. To avoid hand-crafting this parameter on different tasks and demonstrations, we apply a simple annealing strategy on $d_k$ to realize a \textbf{soft} constraint as it can adapt along with the improvement of policy, comparing to a fixed tolerance. Specifically, we adopt the following update rule for $d_k$
\begin{align}
    d_{k+1} \leftarrow d_k + d_k \cdot \epsilon\text{,}
\end{align}
where $\epsilon$ is the annealing factor. We will further demonstrate the advantage on adopting a soft constraint and the strategy on hyper-parameter choosing in our empirical evaluations in Sec.~\ref{sec:abla2}.

\subsection{Solving with Scalable Policy Models}
We've shown the issues of the penalty method for RLfD when the expert data is imperfect, and therefore motivated our new approach that reformulates it as a constrained policy optimization problem~\eqref{equ:final_objective}. Nevertheless, solving it accurately can be rather challenging due to: \textbf{1. Feasibility}, it may be difficult to find a feasible solution with the two constraints. \textbf{2. Scalability}, for policies that are characterized by a model with high-dimensional parameter space, \emph{i.e.} neural networks, the computation cost of the new constraint will become unaffordable. To this end, we propose to approximately solve it by linearizing around $\pi_{\theta_k}$ at each optimization step. Denoting the gradient of the objective as $g$, the current discrepancy at $\theta_k $ as $d_{\theta_k} $ and its gradient as $b$, the Hessian matrix of the KL-divergence as $H$\footnote{The KL constraint should be approximated via second-order expansion since its first order gradient is zero at $\pi_\theta = \pi_{\theta_k}$.}, the linear approximation to~\eqref{equ:final_objective} is
\begin{align}
\begin{split}\label{equ:approx_objective}
    \theta_{k+1} = \argmax_\theta~~& g^{T}(\theta - \theta_k)\\
    \text{s.t.}~~& b^T(\theta - \theta_k) + d_{\theta_k} \leqslant d_k \\
    & \frac{1}{2}(\theta-\theta_k)^TH(\theta-\theta_k) \leqslant \delta\text{.}
\end{split}    
\end{align}
The approximated optimization problem above is convex as $H$ is always positive semi-definite~\cite{trpo}. Therefore, compared to its original form~\eqref{equ:final_objective}, a feasible solution can be found more easily using duality. 
In particular, given $\lambda$ and $\nu$ as the Lagrange multipliers for KL-divergence and discrepancy constraints, a corresponding dual to~\eqref{equ:approx_objective} can be written as
\begin{align}
\begin{split}\label{equ:dual}
    \max_{\lambda \geq 0 \atop \nu \geq 0} -\frac{1}{2\lambda}(g^T u + 2\nu b^T u + \nu^2 b^T r) - \nu c - \lambda\delta\text{,}
\end{split}
\end{align}
where $u = H^{-1}g$, $r = H^{-1}b$, $c=d_k-d_{\theta_k}$. Since the number of variables in this dual problem is much less than the dimension of $\theta$, the computation cost will also be much less than solving~\eqref{equ:final_objective}. A closed-form solution of optimal solution $\lambda^\star$, $\nu^\star$ can be derived by firstly obtaining and substituting $\nu^\star$, then discussing the sub-case and finally gets $\lambda^\star$. Suppose we have the optimal solution $\lambda^\star$, $\nu^\star$ of this dual problem, the solution to the primal one will be
\begin{align}
\begin{split}\label{equ:solution}
    \theta_{k+1}^\star = \theta_k - \frac{1}{\lambda^\star}(u + r\nu^\star)\text{.}
\end{split}
\end{align}
When there is at least one feasible point within the KL constraint (the trust region), we can update the policy parameter $\theta$ by solving the dual for $\lambda^\star$ and $\nu^\star$~\eqref{equ:solution}. However, due to the initialization and approximation error, the proposed update rule may sometimes not satisfy the constraints in~\eqref{equ:final_objective}, especially at the beginning of optimization. In the next section, we will provide more details on ensuring the feasibility.

\begin{algorithm}[t!]
\caption{RLfD with a Soft Constraint}
\begin{algorithmic}\label{alg:main}
\STATE \textbf{Input:} Imperfect expert demonstrations $\mathcal{D}_E=\{\zeta_i^E\}$, initial policy $\pi_{\theta_0}$, initial constraints tolerance $d_0$, $\delta$, annealing factor $\epsilon$, maximal iterations $N$.
\FOR{k = 0 to $N$}
\STATE Sample roll-out $\mathcal{D}_{\pi}$ with $\pi_{\theta_k}$.
\STATE Estimate $\hat{g}$, $\hat{b}$, $\hat{H}$ with samples from $\mathcal{D}_E$ and $\mathcal{D}_\pi$.
\IF{the optimization problem~\eqref{equ:approx_objective} is feasible} 
\STATE {Solve the dual problem~\eqref{equ:dual} to obtain $\lambda^*$, $\nu^*$}.
\STATE {Compute update step proposal $\Delta\theta$ as~\eqref{equ:solution}}.
\STATE {Update the policy by backtracking line-search along $\Delta\theta$ to ensure the satisfaction of constraints}.
\ELSE 
\STATE{Update the policy via the recovery objective~\eqref{equ:recovery}}. 
\ENDIF \\
Annealing the tolerance $d_k$: $d_{k+1} \leftarrow d_k + d_k \cdot \epsilon$.
\ENDFOR\\
\end{algorithmic}
\end{algorithm}

\subsection{Implementation Details}\label{sec:imple}
\noindent
\textbf{The choice of discrepancy measure.} In RLfD, as we can only access the samples (demonstrations) from the expert policy and its occupancy measure, we adopt the non-parametric distance metric MMD~\cite{mmd1,mmd2,mmd3} as the discrepancy measure. The value and gradient \emph{w.r.t.} policy parameters of MMD can be easily computed with demonstrations and agent roll-outs. Moreover, we use the characteristic Gaussian kernel to ensure the following property
\begin{align}
    \operatorname{MMD}[p, q] = 0 \Leftrightarrow p = q\text{,}
\end{align}
where $p$, $q$ denote two distributions. This property can help alleviate the inconsistency between minimizing discrepancy and morphing distributions within the discrepancy constraint and improve the optimization~\cite{mmd_same}.

\noindent
\textbf{Feasibility issue.} The major crux that accounts for the feasibility issue when solving~\eqref{equ:final_objective} can be twofold. One lies in the beginning phase. As the parameter $\theta$ is usually randomly initialized, it may induce infeasibility when the optimization just starts. We propose a recovery strategy that transforms the constraint into an objective to eliminate this issue.
\begin{align}
\begin{split}\label{equ:recovery}
    \theta^\star = \argmin_\theta \mathbb{D}\left[\rho_{\pi_\theta}(s,a) \| \rho_{\pi_{\theta^-}}(s,a)\right]\text{.}
\end{split}
\end{align}
Another source of infeasibility comes from~\eqref{equ:solution}. The update rule may not satisfy the constraints due to the approximation error. To this end, we apply a backtracking line-search along $\Delta\theta = -{\lambda^{\star}}^{-1}(u + r\nu^\star)$ to ensure the constraint satisfaction. To further reduce the computation cost, we also adopt the conjugate gradient method like~\cite{trpo} to approximately compute the inverse of $H$ and its products.

The algorithm detail is summarized in Algorithm~\ref{alg:main}.

\section{Discussion}\label{sec:discuss}
In this section, we will discuss some relevant research on RLfD, and demonstrate how they connect to our method.

\noindent
\textbf{Pre-train with demonstrations.} A straight-forward solution for combining demonstrations in RL will be pre-training agent policy with expert data via imitation learning, \emph{e.g.} behaviour cloning~\cite{lfd1,lfd2}, then proceeding with normal RL~\cite{pretrain1,pretrain2}. The first step is similar to our constrained optimization approach under unsatisfied constraints when the optimization starts, sometimes even have better performance at the beginning. However, this method cannot guarantee the exploration quality in the later RL step; thus the subsequent training can still suffer from poor sample efficiency in the case with large exploration space and sparse feedback.

\noindent
\textbf{Penalty with other discrepancy measures.} There is also some research on investigating different discrepancy measures for RLfD with penalty departures~\cite{ijcaipofd1,pofd}. Notable recent research is \textbf{POfD}~\cite{pofd}, which proposed to leverage Generative Adversarial Networks (GANs)~\cite{GAN} to evaluate the discrepancy between the occupancy measure of expert and agent. In our comparative evaluations, it demonstrates comparative performances than baseline that employs MMD as penalty departures. However, this method requires an extra parameterized model (discriminator) and training procedure (adversarial training), which substantially increase the difficulty of convergence.

\noindent
\textbf{Penalty with annealing.} In Sec.~\ref{sec:soft_constraint}, we've mentioned that our constraint method adopts an annealing strategy to select the constraint factor adaptively. Since our method would expect the optimal policy to stay within the constraint, annealing is more practical than manually specifying a fixed factor for different task and demonstrations. Similarly, this strategy is also applicable to the factor $\lambda$ in penalty method~\eqref{equ:penalty} for suppressing the side effect of imperfect demonstrations. 
However, we should notice that annealing can only partly alleviate this impact before $\lambda$ becomes zero. While in our approach, only the original RL objective is being optimized once the constraint with imperfect expert data is satisfied. In fact, our empirical results in Sec.~\ref{sec:exp_comp} indicate penalty with annealing does perform advantageously than pure penalty method in some evaluated tasks, but there is still a significant gap to our approach using soft constraint. 

\begin{table*}[htbp]
\caption{\label{tab:eval_reward} 
Comparative results (with only 1 imperfect demonstration). All results are measured in the original exact reward.}
\begin{center}
\setlength\tabcolsep{1.2pt}
\begin{tabular}{ccccccc}
\toprule 
 &  \small{MountainCar}  & \small{DoublePendulum} & \small{Hopper}& \small{Walker2d} & \small{HalfCheetah} &\small{Ant} \\
\midrule
$\mathcal{S}$ / $\mathcal{A}$ & $\mathbb{R}^4$ / $\{0, 1\}$ & $\mathbb{R}^{11}$ / $\mathbb{R}^1$ & $\mathbb{R}^{11}$ / $\mathbb{R}^{3}$ & $\mathbb{R}^{17}$ / $\mathbb{R}^{6}$ & $\mathbb{R}^{17}$ / $\mathbb{R}^6$ & $\mathbb{R}^{111}$ / $\mathbb{R}^{8}$ \\
\small{Setting / Demo} & \textbf{S1} / 81.25 & \textbf{S3} / 1488.28 & \textbf{S2} / 969.71 & \textbf{S2} / 1843.75 & \textbf{S2} / 2109.80& \textbf{S2} / 1942.05 \\
\midrule
\small{PPO}& -0.74$\pm$9.61& 302.77$\pm$37.09 & 17.09$\pm$13.54&1.54$\pm$5.75 &978.84$\pm$665.61&-2332.95$\pm$2193.85 \\
\small{MMD-IL} & 82.99$\pm$4.57 &218.43$\pm$13.72 & 118.66$\pm$0.38 & 8.88$\pm$6.07& 161.74$\pm$219.85 &967.83$\pm$0.87 \\
\midrule 
\small{Pre-train} & 83.35$\pm$6.32 & 8928.79$\pm$388.61 & 1356.47$\pm$470.43 & 2607.38$\pm$301.94 & 3831.96$\pm$150.30 &-5377.25$\pm$1682.56 \\
\small{POfD} & 45.01$\pm$28.16 & 628.47$\pm$69.36 & 32.13$\pm$24.23 & -1.48$\pm$0.03 & 2801.59$\pm$66.03 & -68.59$\pm$19.17 \\
\small{Penalty} & -120.29$\pm$48.30 &1902.95$\pm$210.41 &1225.03$\pm$296.52 &286.23$\pm$12.46 & 1517.68$\pm$35.85 &-3711.12$\pm$794.97 \\
\small{Penalty + Ann.} & 79.00$\pm$1.04 & 1671.78$\pm$108.80 & 1220.10$\pm$112.74 &282.00$\pm$6.70 & 2592.94$\pm$870.04 & -116.89$\pm$88.01\\
\small{Ours} & \textbf{83.46$\pm$1.42}& \textbf{9331.40$\pm$5.95} &\textbf{2329.89$\pm$125.85} &\textbf{3483.78$\pm$269.59} &\textbf{4106.69$\pm$95.47}&\textbf{2645.58$\pm$118.55} \\
\bottomrule 
\end{tabular}
\end{center}
\end{table*}

\begin{figure*}[h]
\begin{center}
\includegraphics[width=\linewidth]{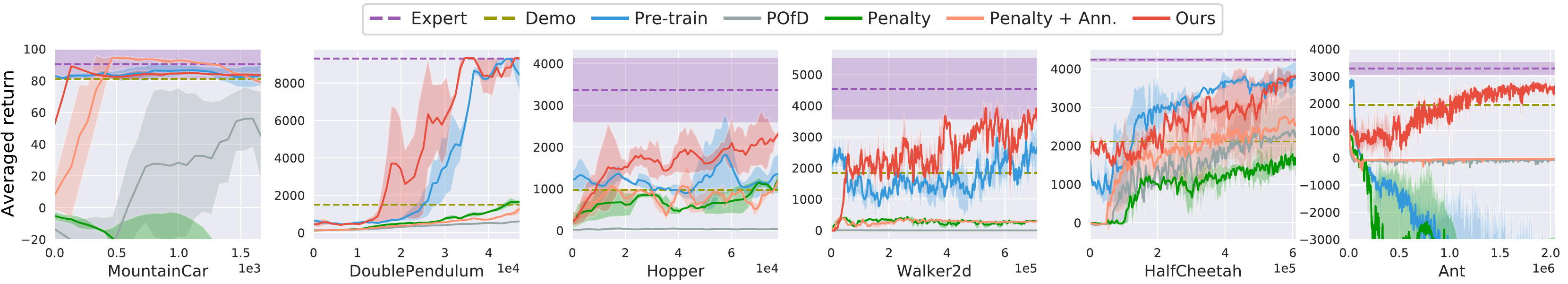}
\caption{Learning curves of our method versus baselines under challenging robotic control benchmark. For each experiment, \textbf{a step represents one interaction with the environment}. The number of steps could be variant in different figures.}
\label{fig:result_compare}
\end{center}
\end{figure*}

\section{Experiments}\label{sec:exp}
For the experiments below, we aim at investigating the following questions:
\begin{enumerate}
    \item Under the same imperfect expert settings, can our method attains better performative results versus the counterparts that do not employ demonstrations as a soft constraint?
    \item How can the different settings of imperfect expert data, \emph{i.e.} quality and amount, affect the performances of our method and baselines?
    \item What is the key ingredient in our method that introduces better empirical results?
\end{enumerate}
To answer the first question, we evaluate our method against several baselines on six physical control benchmarks~\cite{benchmark,gym}, ranging from low-dimensional classical control to challenging high-dimensional continuous robotic control tasks. Regarding the second question, we conduct ablation analysis on the quality and amount of demonstrations, respectively. We test and contrast the performances of our method and two representative baselines (Pre-train~\cite{pretrain1} and POfD~\cite{pofd}) on these different imperfect expert setting. Finally, we explore another ablation analysis on the core component in our method, \emph{i.e.} soft constraint to address the last question.

\subsection{Settings}
To simulate the sparse reward conditions using existing control tasks in Gym, we first propose several reward sparsification methods with details as follows\footnote{The presented results are still evaluated in the original exact reward defined in~\cite{gym}.}:
\begin{itemize}
    \item \textbf{S1}: Agent receives reward $+1$ when it reaches a specific terminal state; otherwise, no reward will be provided.
    \item \textbf{S2}: Agent receives reward $+1$ when has already moved towards a certain direction for some distance.
    \item \textbf{S3}: Agent receives reward $+1$ when its last pole is higher than a given height. Only applied to \textit{DoublePendulum} task.
\end{itemize}

We train expert policies (namely \textit{perfect} experts, shown as \textbf{Expert}) for each tested tasks with PPO~\cite{PPO} based on the exact reward, and select policies learned meanwhile (namely \textit{imperfect} experts, shown as \textbf{Demo}), record only \textbf{one} trajectory as the imperfect demonstrations. To make the comparisons fairer, the policies of all the methods and tasks are parameterized by the same neural network architecture with two hidden layers (300 and 400 units each) and tanh activation functions. All the algorithms are evaluated within the fixed amount of environment steps. And for every single task, we run each algorithm over five times with different random seeds.

\subsection{Comparative Evaluations}\label{sec:exp_comp}
In comparative evaluations, we carry out several RLfD baselines, including Pre-train~\cite{pretrain1} and POfD~\cite{pofd}. In particular, we introduce another two baselines of penalty method\footnote{POfD also belongs to penalty method.} with MMD as discrepancy measure, denoted by Penalty and Penalty + Ann., and the later one also employ an annealing strategy described in Sec.~\ref{sec:discuss}. We also run two non-RLfD baselines PPO and MMD-Imitation (denoted as MMD-IL) to verify the reward sparsification and the imperfect expert setting respectively. PPO will run with the sparse reward while MMD-IL will directly optimize the objective defined in~\eqref{equ:recovery} with provided imperfect demonstrations. In Figure~\ref{fig:result_compare}, the solid curves correspond to the mean reward, and the shaded region represents the variance over five times. The results of our comparative evaluations are summarized in Table~\ref{tab:eval_reward}, which averaged 50 trials under the learned policies.

The results overall read that our method achieves comparable performances with the baselines on relatively simple tasks (such as \textit{MountainCar}) and outperforms them with a large margin on difficult tasks (such as \textit{Hopper}, \textit{Walker2d} and \textit{Ant}). During policy optimization, our method can converge faster than other RLfD counterparts as well as obtains better final results. Comparing with the strong baseline of Pre-train, we can see that although convergence efficiency of proposed method during the early phase of training may not have significant advantages, but as it continues, the performance of our method can be improved persistently like \textit{Hopper}(+973.42) and \textit{Walker2d}(+876.40), while Pre-train struggles on achieving higher return, which demonstrates that our method could benefit more from the exploration guidance offered by the soft constraint during the whole policy optimization procedure than by only imitating at the beginning.

On the other hand, we also find that our algorithm exhibits a more stable and efficient behavior over all the baselines using the penalty method. From the learning curve and numerical results, it can be seen that adopting penalty with imperfect demonstrations will induce a noisy and misleading gradient update, which will prevent the performances from improving further while our method with a soft constraint will not suffer from this. This essentially accounts for the performance gap between our method and all baselines with penalty departures. Moreover, the complex training strategies and auxiliary model in POfD also leads to unstable and inefficient training across different tasks and environment specifications. 

From the results of PPO and MMD-Imitation, the experiment settings of reward sparsification and imperfect demonstrations can be verified. As it illustrates, under sparse environmental feedback, pure PPO fails to find an optimal policy on most of the tested tasks, which indicates the impact of ineffective exploration. While with few imperfect demonstrations, MMD-Imitation also cannot learn promising policies. It suggests that combining the demonstrations and environmental feedback would be essential for the designated tasks. Furthermore, as similar MMD-Imitation update may happen in our method when the optimization just starts (mentioned in Sec.~\ref{sec:imple}), these results also show how can our method benefit from the follow-up solving of the constrained optimization problem.

\subsection{Ablation Analysis I: Sensitivity to Demonstrations}
The results presented in the previous section suggest that our proposed method outperforms other RLfD approaches on several challenging tasks. We're now interested in whether these advantages still hold when the demonstration setting changes. We will compare our method and baselines on demonstrations with different amounts and quality respectively to show how can they affect the performative results.

\begin{figure}[t!]
\begin{center}
\centering
\includegraphics[width=.91\linewidth]{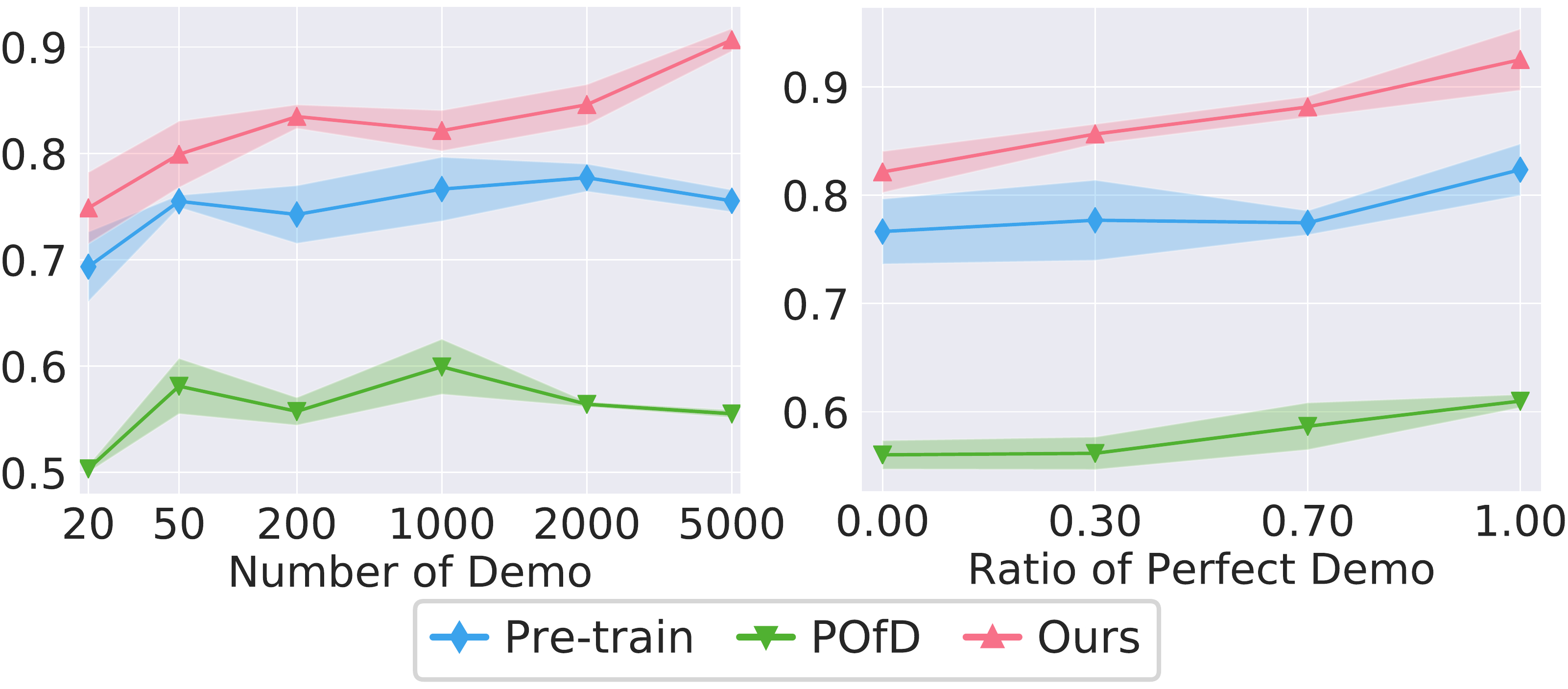}
\caption{Results on \textit{HalfCheetah} task with different imperfect expert setting. \textbf{Left}: Different number of state-action pairs; \textbf{Right}: Different level of imperfectness.}
\label{fig:result_diff_demo}
\end{center}
\end{figure}

\noindent
\textbf{Demonstrations with different amounts.} We select six groups of demonstrations with different amounts from 50 to 5000 for comparison on the \textit{HalfCheetah} task. Notice the comparative experiments in Sec.\ref{sec:exp_comp} are conducted with one trajectory with 1000 state-action pairs as demonstrations. The corresponding results are plotted in the \textbf{Left} of Figure~\ref{fig:result_diff_demo}. The results read that our method performs advantageously than the baselines on these demonstration settings, and the performance gap is also getting larger as the number of demonstrations increases. On the other hand, the results could benefit from more demonstrations in a certain range for all the methods, while our method can be more robust when the demonstrations become fewer.

\noindent
\textbf{Demonstrations with different qualities.} We emulate the demonstrations of different qualities by mixing the demonstrated data from perfect (\textbf{Expert}) and imperfect (\textbf{Demo}) policies with different ratios. The \textbf{Right} of Figure~\ref{fig:result_diff_demo} presents the results of our method and baselines with these demonstrations. It implies that the quality of demonstrations will significantly affect the performances of all the evaluated methods, and expert data with high quality can facilitate policy optimization to a certain extent. We can also see that our method overall outperforms the two counterparts even though the expert data becomes perfect (by setting the ratio to 1.00), indicating that our constraint-based method can exploit the expert data more efficiently than other methods based on penalty departures or pre-training.

\subsection{Ablation Analysis II: Sensitivity to Constraint Tolerance}\label{sec:abla2}
Now we will further investigate how can the design of the core soft constraint affect the performative results of our method. More specifically, we're interested in the tolerance factor $d$. By varying the initial value of $d$ and annealing strategies (namely, different annealing factor $\epsilon$), we will explore the sensitivity of our algorithm regarding them.

\begin{figure}[t!]
\begin{center}
\centering
\includegraphics[width=1.\linewidth]{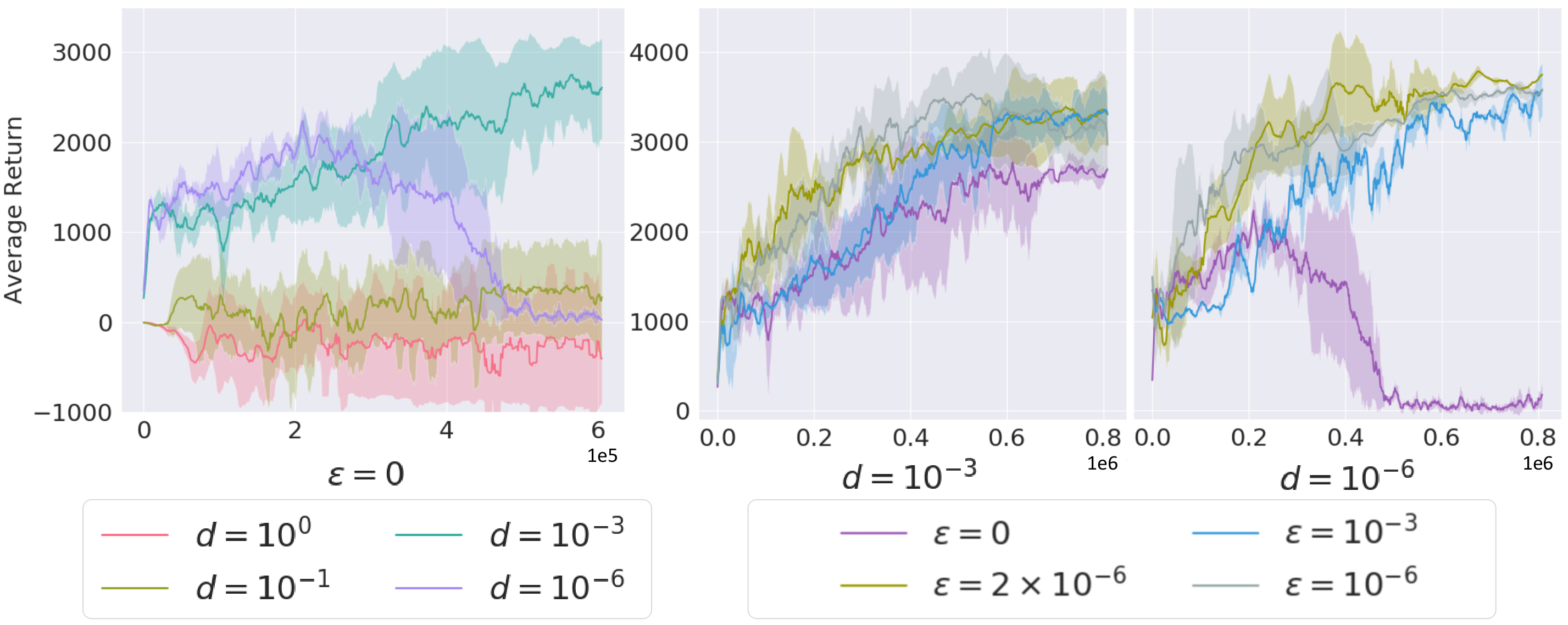}
\caption{Learning curves over on \textit{HalfCheetah} task. \textbf{Left}: ablation study about different tolerance factor $d$; \textbf{Right}: sensitivity of choosing fixed or annealing strategy of tolerance.}
\label{fig:result_ablation}
\end{center}
\end{figure}

\noindent
\textbf{Different tolerance.} We design four groups of parameters for the ablation experiments on the tolerance choosing in \textit{HalfCheetah} task, where the annealing mechanism is disabled by setting $\epsilon$ fixed at zero, and choose initial tolerance $d_0$ from $\{10^0,10^{-1},10^{-3},10^{-6}\}$. The learning curves are plotted in \textbf{Left} of Figure~\ref{fig:result_ablation}. As the results demonstrate, when given relatively large tolerance, the exploration reference from demonstrations will not work as the constraint almost does not affect policy optimization. In contrast, a too-small tolerance will hurt the final performance when the demonstrations are imperfect. Therefore, hand-crafting the tolerance for the constraint can be difficult, and an automatic adjustment with the annealing mechanism should be adopted. 

\noindent
\textbf{Fixed vs. Annealing tolerance.} In the previous experiment, we mention the importance of annealing of tolerance. Now we explore the advantages of annealing mechanism quantitatively in \textit{HalfCheetah} task. Since our annealing is to enlarge the tolerance along training, we only choose two not-too-large initial tolerances $d_0$ from $\{10^{-3}, 10^{-6}\}$, and select the annealing factor $\epsilon$ from $\{0,2\times10^{-3},10^{-3}, 10^{-6}\}$. Corresponding learning curves are shown in \textbf{Right} of Figure~\ref{fig:result_ablation}. We can see that the performances of our method with an annealing tolerance are overall better than with a fixed one (simply by setting $\epsilon$ as zero). Moreover, when the annealing factor $\epsilon$ is set properly, the performance of our method is not sensitive to the minor changes of $\epsilon$ as the results of different factors are almost at the same level. This further demonstrates the robustness of our proposed method.

\section{Conclusion}\label{sec:conclusion}
In this paper, we investigate the problem of RLfD with imperfect expert data. Compared to existing RLfD problem setting, this new setting does not require the expert to be optimal, which can be more practical for real-world demonstrators like a human. We show that current penalty based RLfD methods will suffer from the issue of optimality and convergence when being applied to the setting of imperfect experts both theoretically and empirically. To this end, we propose to employs the expert data as a soft constraint and reformulate RLfD as a constrained policy optimization problem to narrow the negative impact of the imperfectness. We also provide an efficient learning algorithm for solving the challenging constrained optimization problem with scalable policy model like neural networks. Experiments on physical control benchmarks demonstrate the effectiveness of our proposed method over other RLfD counterparts. While we still assume the expert data to be collected from the same domain as the current conducted task, further exploration on combining our work with representation learning to enable learning with demonstrations across different domains could be a new direction of future work.

\section*{Acknowledgment}
This research was funded by National Science Foundation of China (Grant No.91848206). It was also partially supported by the National Science Foundation of China (NSFC) and the German Research Foundation (DFG) in project Cross Modal Learning, NSFC 61621136008/DFG TRR-169. We would like to thank Dr.~Boqing Gong and Dr.~Tao Kong for their generous help and insightful advice.

{\fontsize{9pt}{10pt} \selectfont 
\bibliographystyle{aaai}
\bibliography{bibliography}
}
\end{document}